\documentclass[preprint]{elsarticle}
\usepackage{graphicx}
\usepackage{natbib}
\usepackage{amsmath}
\usepackage{multirow}
\usepackage{lscape}
\usepackage{xcolor}
\usepackage{array, makecell}
\usepackage{subcaption}
\usepackage{svg}
\usepackage{adjustbox}
\usepackage{natbib}
\usepackage{hyperref}

\journal{Ecological Informatics}

\begin{document}

\begin{frontmatter}

\title{\textcolor{black}{Linear probing enables Ship-Radiated Noise recognition with pretrained audio embeddings}}
\author[inst1]{Hilde I. Hummel*}
\ead{h.i.hummel@cwi.nl}
\affiliation[inst1]{organization={Centre of Mathematics and Computer Science, Stochastics group},
            addressline={Science Park 123}, 
            city={Amsterdam},
            postcode={1098 XG},
            country={Netherlands}}

\author[inst2]{Sandjai Bhulai}
\author[inst1,inst2]{Rob van der Mei}
\author[inst3]{Burooj Ghani}

\affiliation[inst2]{organization={Vrije Universiteit, Department of Mathematics},
            addressline={De Boelelaan 1111}, 
            city={Amsterdam},
            postcode={1081 HV},
            country={Netherlands}}

\affiliation[inst3]{organization={Naturalis Biodiversity Center},
            addressline={Darwinweg 2}, 
            city={Leiden},
            postcode={2333 CR},
            country={Netherlands}}

\cortext[inst1]{Corresponding author}

\begin{abstract}
Even though the ocean covers the majority of the planet's surface, it remains the least explored ecosystem. As light and radio waves do not propagate through water, underwater acoustics is the main choice for various ocean applications ranging from marine biology to pollution monitoring. Increasing levels of anthropogenic noise from ships contribute significantly to underwater sound pollution, posing risks to marine ecosystems. This makes monitoring crucial to understand and quantify the impact of the ship radiated noise. Passive Acoustic Monitoring (PAM) systems are widely deployed for this purpose, generating years of underwater recordings across diverse soundscapes. Manual analysis of such large-scale data is impractical, motivating the need for automated approaches based on machine learning. Recent advances in automatic Underwater Acoustic Target Recognition (UATR) have largely relied on supervised learning, which is constrained by the scarcity of labeled data. Transfer Learning (TL) offers a promising alternative to mitigate this limitation. In this work, we conduct the first empirical comparative study of transfer learning for UATR, evaluating multiple pretrained audio models originating from diverse audio domains. The pretrained model weights are frozen, and the resulting embeddings are analyzed through classification, clustering, and similarity-based evaluations. The analysis shows that the geometrical structure of the embedding space is largely dominated by recording-specific characteristics. However, a simple linear probe can effectively suppress this recording-specific information and isolate ship-type features from these embeddings. As a result, linear probing enables effective automatic UATR using pretrained audio models at low computational cost, significantly reducing the need for a large amounts of high-quality labeled ship recordings. 

\end{abstract}




\begin{keyword}
Ship Radiated Noise \sep Pretrained Audio Models \sep UATR \sep PAM \sep Underwater Acoustics
\PACS 0000 \sep 1111
\MSC 0000 \sep 1111
\end{keyword}

\end{frontmatter}

\section{Introduction}
The ocean covers roughly 71\% of the planet's surface, however, it is the least explored ecosystem  \cite{ramirez2010deep}. As light and radio waves do not propagate well through water, exploration of the ocean is mainly driven by acoustic analysis \cite{li2025underwater}. This leads to diverse applications of underwater acoustics ranging from climate change measurement and marine biology to pollution monitoring, such as sound pollution \cite{li2025underwater}. The latter comes from various anthropogenic underwater noises, such as\textit{ Ship-Radiated Noise} (SRN). Due to the increase in activity of ships over the years, the increased levels of anthropogenic noise generated by ships pose a threat to the sustainability of marine ecosystems \cite{williams2015impacts, chahouri2022recent}. The low-frequency sounds produced by the engines, propellers, and hull vibrations of ships \cite{hummel2024survey} interfere with the natural acoustic environment of the oceans, disrupting vital behaviors such as communication between marine species \cite{williams2015impacts}. This makes monitoring of ship sounds crucial to better understand and mitigate their ecological impact \cite{hummel2024survey}. 

For this, Passive Acoustic Monitoring (PAM) systems are deployed to study the underwater soundscape. PAM systems enable continuous, non-invasive observation of ocean environments by recording ambient sounds through hydrophones. Since PAM systems do not emit an acoustical signal but only listen, their monitoring does not disturb the marine ecosystem. This makes PAM particularly suitable for long-term ecological studies and for detecting anthropogenic noise in various habitats. 

Although PAM supports research on marine mammal detection \cite{bittle2013review}, fish biodiversity \cite{lindseth2018underwater}, and even climate change \cite{affatati2022ocean}, this study focuses on SRN. Each individual ship has its own unique sound profile, which can be used to identify its type and even the individual ship \cite{hummel2024survey}. This process is referred to as \textit{Underwater Acoustic Target Recognition} (UATR). Manual analysis of these recordings is time-consuming and labor-intensive. Given that PAM systems often produce years of unlabeled continuous recordings, the resulting data volume makes manual annotation and analysis impractical. This creates the need for the automation of this process. 

To this end, a variety of machine learning (ML) methods have been suggested to automatically monitor ship sounds \cite{hummel2024survey}. Most of these studies focus on supervised learning, which limits their use due to the scarcity of labeled data. The labeled datasets span a small region of the ocean \cite{santos2016shipsear, irfan2021deepship}, which limits the environmental diversity within the recordings. In addition, due to the small size of the dataset, it is not feasible to train large ML models. Therefore, these methods may not be generalizable to newly seen ships or other ocean environments \cite{hummel2025computation}. To cope with the limited amount of labeled data, \textit{Transfer Learning} (TL) has emerged as a promising approach, as demonstrated in fields such as computer vision \cite{kornblith2019better, mensink2021factors}, general audio processing \cite{tsalera2021comparison}, and bioacoustics \cite{ccoban2020transfer, ghani2025impact}. 

In TL, the backbone of the model is first pretrained using a large dataset of a related domain. By doing so, it has learned some general features that could be generalizable to other domains. Next, the pretrained model can be repurposed as a feature extractor to extract information from another dataset. These features can then be applied to perform a new task, where only a small classifier needs to be trained instead of the whole model \cite{ghani2025impact}. This approach reduces the number of trainable parameters and, therefore, a small labeled dataset is sufficient, eliminating the need for a large qualitative labeled dataset \cite{van2024birds}. However, the success of TL depends on the choice of the backbone of the pretrained model \cite{schwinger2025foundation}. Numerous pretrained models based on large-scale audio data have recently been proposed \cite{zaman2023survey}, and several studies have shown their effectiveness in various downstream tasks \cite{turian2022hear, saeed2021contrastive}. 

\begin{figure}
    \centering
    \includegraphics[width=0.9\linewidth]{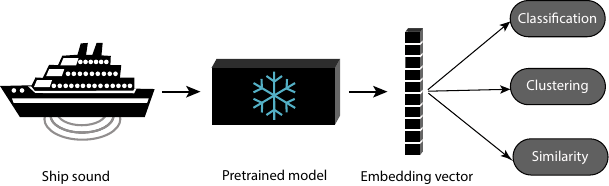}
    \caption{Illustration of the evaluation method for various pretrained audio models.}
    \label{fig:overview}
\end{figure}

A common approach involves freezing the pretrained model weights and using the resulting embedding vectors as input features for a simple classifier \cite{rauch2025can, kather2025clustering}. To the best of our knowledge, no comprehensive study has examined the transferability of these pretrained audio models to UATR. Motivated by this, in this work, a range of pretrained audio models from various audio domains are systematically evaluated to assess their generalizability to UATR. The performance of the models is evaluated using three complementary methods in two benchmark datasets: (1) ship-type classification through linear probing, (2) clustering-based embedding analysis, and (3) similarity-based embedding evaluation. Together, these methods show classification performance and the representational quality of the embedding vectors generated. An illustration of this is given in Figure \ref{fig:overview}. All data, code, and trained models used in this study are openly available under an Open Access license at this \href{https://github.com/hildeingvildhummel/Decodable_not_Structured}{GitHub page}\footnote{\url{https://github.com/hildeingvildhummel/Decodable_not_Structured}}
to ensure transparency and reproducibility. In summary, the contributions of this study are as follows:

\begin{itemize}
    \item A systematic review of a number of pretrained models in various audio domains;
    \item An extensive empirical study on the transfer of these models to automatic UATR;
    \item A combination of evaluation metrics to evaluate and compare the performance of pretrained audio models for UATR.
\end{itemize}

\section{Related work}
This section describes recent work on automated UATR and briefly describes comparison studies for other audio domains. 

\subsection{Classification of ships}
Previous work on automated UATR has primarily focused on \textit{Supervised Learning} (SL) \cite{hummel2024survey}. These methods have largely relied on \textit{Convolutional Neural Networks} (CNNs), although more recent work has shifted the focus to Transformer-based models. However, limited attention has been paid to the generalizability of these models \cite{muller2024navigating}. This gap has been addressed in \cite{hummel2025computation}, where a small Conformer model is proposed that is optimized using \textit{Self-Supervised Learning} (SSL). The model is trained on a large amount of unlabeled data extracted from a single hydrophone. After pretraining, the model weights were frozen, and a linear classifier was added for downstream classification. The model was shown to be generalizable in both the classification of ship type and the classification of marine mammal calls. Recently, SSL frameworks have received increasing attention for UATR. 

For example, \cite{xu2023self} proposed an SSL framework in which the pretraining of the framework has not been done on underwater acoustic data, but they utilized a general dataset called AudioSet \cite{gemmeke2017audio}. The framework employs a so-called Swin Transformer encoder that processes masked and patched Mel-spectrograms, with one decoder reconstructing the masked patches and the other reconstructing the gammatone spectrogram of the original audio. Here, the first decoder captures local information, and the second focuses on the global information of the recording. They evaluated their method on the benchmark dataset Deepship, reaching 80\% accuracy \cite{gemmeke2017audio}. Similarly, \cite{li2025cross} also performed the pretraining of their learning framework on AudioSet. Their framework consists of two modules, a combination of masked learning and contrastive learning using a Swin Transformer during pretraining. The knowledge captured within the Transformer model is then transferred to a light-weight CNN model. This reduces the number of parameters, facilitating the possibility of using the model on edge devices. Their model achieved over 87\% accuracy on the Deepship dataset. Although these studies introduce novel architectures and training paradigms, none of the well-established SSL methods have been proven to be effective in other audio domains and have yet been systematically applied to UATR.

\subsection{Previous comparison studies}
For speech-based models, a standardized benchmark known as the Speech Processing Universal PERformance Benchmark (SUPERB) \cite{yang2021superb} has been proposed to fairly evaluate the generalizability of models across a wide range of speech-related tasks. This approach is extended in \cite{yang2024large}, where they evaluated 33 speech foundation models in 15 speech-related tasks. These tasks range from speaker verification to speech enhancement to automatic speech recognition and keyword recognition. Each layer of every model was evaluated, motivated by the understanding that different layers of \textit{Bidirectional Encoder Representations from Transformers} (BERT) capture distinct types of information. Their findings demonstrated that SSL models have superior performance in the generalizability to these various speech-related tasks. However, these speech models have thus far been tested only on speech-related downstream tasks and do not take any other audio domain into account. 

In parallel, various studies have been conducted to explore the potential of TL in the bioacoustic domain. Kather \textit{et al.} \cite{kather2025clustering} compared 15 bioacoustic models by classifying either birds or frogs using these pretrained bioacoustic models. In addition, they evaluated the generated embedding spaces by applying clustering and Adjusted Mutual Information \cite{kather2025clustering}. In this study, they concluded that SL models still outperform SSL ones overall, though SSL models exhibit better generalization to new tasks such as frog classification. 

Schwinger \textit{et al.} \cite{schwinger2025foundation} compared bioacoustic and general audio models, providing an extensive review of their architectures and evaluating their performance on two bioacoustic datasets using both linear and attentive probing. Here, they showed that Transformer models benefit from attentive probing, achieving performance comparable to or exceeding that of SL models. Similarly, Miron \textit{et al.} \cite{miron2025matters} compared various general audio models with several bioacoustic models. They extended the research \cite{kather2025clustering} by using clustering to evaluate the embedding space and adding a retrieval evaluation. During retrieval evaluation, the embedding space is directly evaluated using the cosine similarity of each test sample compared to the full test set. They found that pretraining on a mixture of bioacoustic and general audio data, followed by supervised post-training on the same mixture, yields the best in- and out-of-distribution performance.

\section{Methods}
This section outlines the datasets used for pretraining and benchmarking, followed by descriptions of the selected models and evaluation methodologies.

\subsection{Datasets}
The models employed in this study were pretrained on various domain-specific datasets prior to further evaluation.
Subsequently, the performance of the models are evaluated using two benchmark datasets. This section provides a detailed description of both the pretraining and benchmark datasets.

\subsubsection{Datasets: pretraining datasets}
Depending on the audio domain, various pretraining datasets are in place. A well-known dataset for general audio purposes is AudioSet \cite{gemmeke2017audio}, which contains more than 2 million annotated 10-second audio files sourced from YouTube. This makes the dataset highly diverse, containing more than 600 classes, although the content is still dominated by music and speech. It can be difficult to obtain the complete AudioSet, because YouTube videos may be removed over time. To address this, another dataset, FSDK50 \cite{fonseca2021fsd50k} has been introduced. This dataset contains over 51k audio clips that were manually annotated and derived from the AudioSet ontology. Similarly, VGGSound \cite{chen2020vggsound} extracted audio and visuals from YouTube videos, resulting in a dataset of paired audio-visual samples. In the speech domain, Librispeech \cite{panayotov2015librispeech} is a large dataset considering English-only speech. This dataset was explicitly designed to train automatic speech recognition systems. Unlike general audio datasets, this dataset does not contain classes, but transcriptions of the speech that have been recorded. Beyond speech and general audio, the availability of bioacoustic datasets has grown to monitor biodiversity in a non-invasive manner. Bird sound datasets, in particular, have reached a substantial scale. In this domain, Xeno-Canto \cite{vellinga2015xeno} plays a central role in this domain. This dataset offers globally observed bird songs that have been recorded by various devices, spanning more than 10,000 classes. Based on Xeno-Canto, another large-scale labeled dataset BirdSet is introduced \cite{rauch2024birdset}. This dataset is a collection of several bioacoustic datasets, with the focals derived from the Xeno-Canto dataset. 

In addition to bird recordings, other species are also represented in recent bioacoustic efforts. The Meerkat data set \cite{schafer2024animal2vec} comprises more than 1,000 hours of annotated meerkat vocalizations labeled by type of social interaction. In total, the recordings are 10 seconds long, completing a total of 1,068 hours of recordings. The final bioacoustic dataset taken into account is the iNatSounds dataset \cite{chasmai2024inaturalist}. This dataset differentiates itself from the others by incorporating sounds from a larger diversity of animals than only birds or meerkats, including mammals, insects, reptiles, and amphibians, resulting in more than 5,500 species. Unfortunately, all of these recordings are weakly labeled. This means that it contains a single positive label for the entire recording. Therefore, it is possible that the majority of the recording contains background noise or is even interrupted by the sound of another species. Aquatic ecosystems are also being studied through passive acoustic monitoring. To automatically monitor biodiversity, especially within coral reefs, a dataset ReefSet \cite{williams2024leveraging} is introduced. This dataset is a combination of 16 individual datasets that have been recorded across 12 countries. Additionally, \textit{National Oceanic and Atmospheric Administration} (NOAA) \cite{noaa2017passive} has an extensive amount of unlabeled data that can be utilized for underwater biodiversity monitoring. In \cite{googlewhale}, they annotated part of the NOAA data themselves to train the model. A complete overview of all the pretraining datasets is given in Table \ref{tab:datasets}.


\begin{table}[]
    \centering
    \begin{tabular}{l|l|r|c|c|c|c}
        Type & Dataset &  Duration & Labeled & \# Classes & Descriptor & Citation\\
        \hline
        \multirow{3}{*}{General Audio} & AudioSet & 5,790 hrs & Yes & 631 & AS & \cite{gemmeke2017audio}\\
         & FDSK50 & 100$+$ hrs & Yes & 200 & FDSK50 & \cite{fonseca2021fsd50k} \\
         & VGGSound & 550 hrs & Yes & 309 & VGG & \cite{chen2020vggsound} \\
         \hline
        \multirow{1}{*}{Speech} & Librispeech & 960 hrs & Yes & Transcriptions & LS & \cite{panayotov2015librispeech} \\
        \hline
        \multirow{4}{*}{Bioacoustics} & Xeno-Canto & $\pm$ 1,100 hrs & Yes & 10,000$+$ & XC & \cite{vellinga2015xeno} \\
        & BirdSet & 6,877 hrs & Yes & $\pm$10,000 & BS & \cite{rauch2024birdset}\\
        & iNatSounds & $\pm$ 1,200 hrs & Yes & 5,500$+$ & iN & \cite{chasmai2024inaturalist} \\
        & MeerKAT & 1,068 hrs & Partially & 8 & MK & \cite{schafer2024animal2vec}\\
        \hline
        \multirow{4}{*}{Marine Life sounds} & Reefset & $\pm$1,800 hrs & Yes & 37 & RS & \cite{williams2024leveraging}\\
        & \makecell{National Oceanic \\ and Atmospheric \\ Administration} & $-$ & No & $-$ & NOAA & \cite{noaa2017passive}\\
    \end{tabular}
    \caption{Overiew of pretraining datasets.}
    \label{tab:datasets}
\end{table}

\subsubsection{Benchmark datasets: The evaluation sets}
For the evaluation of the proposed models, two benchmark datasets were selected. The first, Deepship \cite{irfan2021deepship}, contains more than 45 hours of single-ship recordings collected in the water near Vancouver. The annotations were made by defining an inclusion zone of a two km range around the hydrophone, such that each single ship within this range was defined as recorded. The dataset comprises four ship classes. 

The second benchmark dataset, ShipsEar \cite{santos2016shipsear}, is smaller compared to Deepship, totaling eight hours of single-ship recordings. The audio was recorded on the Spanish Atlantic coast in northwest Spain and consists of five classes. Both Deepship and ShipsEar are commonly used datasets for the evaluation of automatic UATR \cite{hummel2024survey}. Both datasets were divided into a training set and a test set. For Deepship, a time-wise split was made as described in \cite{hummel2025computation}. This split ensures that there is no data leakage in the test set and that the test set represents the future relative to the training set. For ShipsEar, this information is not present to make such a split. Therefore, 80\% of the individual recordings were utilized for training, and the remaining recordings for testing. Both datasets represent distinct acoustic environments and vessel populations, making cross-dataset evaluation a useful measure of model generalizability.


\subsection{Models: Pre-trained models used in this study}
The models in this comparative study were selected based on the audio domain in which they were initially trained. This could either be ''\textit{general audio}'', ''\textit{speech}'', ''\textit{bioacoustics}'', or ''\textit{marine life sounds}''. Models trained on a large amount of data and possessing the potential to generalize well to other related audio domains were selected. In addition, these models are not pretrained on either of the benchmark datasets, and they need to have a clearly defined embedding space for evaluation purposes. This is important, since it is then clear where the information bottleneck is defined within the architecture, and therefore, it is clear which layer should be used to extract the features. An overview of the selected models is given in Table \ref{tab:models}. For some of these models, multiple sizes are available. If this were the case, the base models were chosen for evaluation. 

\begin{landscape}
\begin{table}[]
    \centering
    \renewcommand{\arraystretch}{1.2}
    \begin{adjustbox}{max width=\linewidth, max height=\textheight}
    {\begin{tabular}{l|l|l|c|c|l|c|c|c|c|c|c}
        Domain & Model & Architecture & Num. params & Dataset & Input type & Samplerate & Window size & Embedding size & Training paradigm & Open Access & Citation\\
        \hline
         \multirow{ 2}{*}{General Audio} & AudioMAE & Transformer & 86M & AS & Mel-spectrogram & 16 kHz & 10 sec & 768 & SSL & Yes & \cite{huang2022masked}\\
         & BEATS & Transformer & 90M & AS & Spectrogram & 16 kHz & 10 secs & 768 & SSL & Yes & \cite{chen2022beats}\\
         \hline
         \multirow{ 4}{*}{Speech} &  Wav2Vec2.0 & Transformer & 95M &  LS & Raw audio & 16 kHz & 10 sec & 768 & SSL & Yes & \cite{baevski2020wav2vec}\\
         & Data2Vec & Transformer & 95M & LS & Raw audio & 16 kHz & 10 secs & 768 & SSL & Yes & \cite{baevski2022data2vec}\\
         & WavLM & CNN and Transformer & 94.70M & LS & Raw audio & 16 kHz & 10 secs & 768 & SSL & Yes & \cite{chen2022wavlm} \\
         & HuBERT Base & CNN and Transformer & 95M & LS & Raw audio & 16 kHz & 10 secs & 768 & SSL & Yes & \cite{hsu2021hubert}\\
         \hline
         \multirow{ 7}{*}{Bioacoustics} & BirdMAE & Transformer & 86M & BirdSet & Spectrogram & 32 kHz & 10 sec & 1,280 & SSL & Yes & \cite{rauch2025can}\\
         & BirdNet & ResNet & 27M & XC, eBird, Macauley Library & Spectrogram & 48 kHz & 3 secs & 1,024 & SL & Yes & \cite{kahl2021birdnet}\\
         & AvesEcho & PaSST & 85M & - & Mel-spectrogram & 16 kHz & 1 sec & 768 & SL & Yes & \cite{ghani2024generalization}\\
         & Animal2vec MK & Transformer & 315M & MK & Raw audio & 8 kHz & 10 secs & 1,024 & SSL & Yes & \cite{schafer2024animal2vec}\\
        & Perch & EfficientNet & 7.8M & XC & Raw audio & 32 kHz & 5 secs & 1,280 & SL & Yes & \cite{ghani2023global}\\
        & Perch 2.0 & EfficientNet & 12M & XC, iN,
        Tierstimmenarchiv, FSDK50 & log Mel-spectrogram & 32 kHz & 5 sec &  1,536 & SL & Yes & \cite{van2025perch}\\
        & AVES & CNN and Transformer & 95M & FSD50K, AS 20k, VGG & Raw audio & 16 kHz & 1 sec & 768 & SSL & Yes & \cite{hagiwara2023aves}\\
        \hline
        \multirow{ 2}{*}{Marine life sounds} & Google Whale & EfficientNet & & NOAA & Spectrogram & 24kHz & 5 sec & 1,280 & SL & Yes & \cite{googlewhale}\\
        & SurfPerch & EfficientNet & & RS & Spectrogram & 32 kHz & 5 sec & 1,280 & SL & Yes & \cite{williams2025using}\\
        
    \end{tabular}}
    \end{adjustbox}
    \caption{\textcolor{black}{Information about the pretrained models chosen for this comparison study.}}
    \label{tab:models}
\end{table}
\end{landscape}

\subsubsection{General audio models}\label{sec:GeneralAudioModels}
For the general audio domain, three individual pretrained models were selected: (1) Audio Masked AutoEncoder (AudioMAE) \cite{huang2022masked}, (2) Bidirectional Encoder representation from Audio Transformers (BEATS) \cite{chen2022beats}, and (3) \textit{Hidden-Unit BERT AudioSet} (HuBERT AS) \cite{ARCH}. Each model employs a distinct self-supervised learning strategy. AudioMAE first converts the raw audio into a Mel-spectrogram. These spectrograms are then divided into patches, where some of these patches are masked before being passed to the model. The masked patches are encoded using a Transformer-based encoder, and the resulting representations are processed by a Transformer-based decoder with local attention mechanisms. The decoder tries to reconstruct the masked patches of the input signal and is trained using a simple Mean Squared Error (MSE). This reconstruction objective primarily focuses on the correctness of low-level time-frequency features, but does not take into account the semantic abstraction of high-level audio \cite{ramesh2021zero, bao2021beit}. To include this in the training process, BEATS introduced discrete label prediction in the general audio domain. The model employs an acoustic tokenizer to generate discrete labels from unlabeled audio. The SSL framework is then optimized jointly using both a masked reconstruction loss and a discrete label prediction loss, enabling BEATS to capture both acoustic and semantic information. Similarly, HuBERT \cite{hsu2021hubert} uses pseudo-labels to train its representations. During the first iteration, these labels are generated by clustering MFCC features using a simple $K$-Means algorithm. In the subsequent iterations, a random layer from the encoder is selected, and its outputs are clustered to form progressively more informative pseudo-labels. Although originally designed for speech representation learning, for this work, a HuBERT model trained on the full AudioSet is selected \cite{ARCH}. This version is referred to as HuBERT AS. 

\subsubsection{Speech models}
In recent years, several pre-trained speech models have been proposed \cite{malik2021automatic}. In this study, four representative models were selected: Wav2Vec 2.0 \cite{baevski2020wav2vec}, Data2Vec \cite{baevski2022data2vec}, WavLM \cite{chen2022wavlm}, and HuBERT \cite{hsu2021hubert}. Wav2Vec2.0 \cite{baevski2020wav2vec} consists of a CNN encoder that inputs the raw audio. The output of this encoder is then fed to a Transformer to extract the representation. In addition, the output of the CNN encoder is discretized using a quantization module. During pretraining, a contrastive learning objective is employed, portions of the encoder output are masked before being passed to the Transformer, while the full latent sequence is quantized. The model learns to identify the correct quantized latent audio representation. This setup has shown great performance in speech-related tasks \cite{kunevsova2024comparison}. Therefore, a similar approach is utilized in Data2Vec \cite{baevski2022data2vec}. Here, they presented a general framework for SSL that can be applied to vision, text, and speech. For speech, Data2Vec adopts a similar architecture to Wav2Vec 2.0 but replaces the contrastive loss with a latent reconstruction objective based on knowledge distillation. Another approach is presented in WavLM \cite{chen2022wavlm}. Again, this model consists of a CNN encoder coupled to a Transformer. However, they suggested Gated Relative Position Bias \cite{chi2021xlm} in the Transformer encoder. The model is pretrained using a masked speech denoising and prediction objective. They suggest that the combination of this objective improves robustness for complex environments \cite{chen2022wavlm}, making WavLM a promising candidate for transfer learning to other audio domains. Finally, HuBERT \cite{hsu2021hubert}, as described in Section~\ref{sec:GeneralAudioModels} above, learns through pseudo-label prediction. It generates labels by clustering features extracted from the raw audio and iteratively refines these labels across training stages. Although originally designed for speech representation learning, its structure allows it to generalize well to other audio-based applications.

\subsubsection{Bioacoustic models}
Beyond the general audio and speech domain, the bioacoustic domain presents an equally valuable field for evaluation. This domain has access to an enormous amount of labeled data \cite{vellinga2015xeno, chasmai2024inaturalist}, allowing supervised training of large models. Such large models are expected to generalize to similar audio applications. Moreover, the implementation of various pretrained bioacoustic models is facilitated in the repository called \textbf{bacpipe} developed by Kather \textit{et al.} \cite{kather2025clustering}. This repository provides standardized tools for model evaluation and comparison. In this study, several representative bioacoustic models were selected for evaluation. 

The first, BirdMAE \cite{rauch2025can}, is inspired by the original AudioMAE \cite{huang2022masked} and trained on a large number of bird songs. During pre-training, they suggested some small changes; for instance, they replaced the original Swin Transformer with a Vision Transformer. In addition to BirdMAE, another popular bird model, called BirdNet \cite{kahl2021birdnet}, has been selected. This CNN-based model is optimized in a supervised manner and is able to classify almost 10,000 different bird species. As a follow up on BirdNET, AvesEcho is developed \cite{ghani2024generalization}, where a student model is optimized with BirdNET as the teacher using knowledge distillation.

Beyond avian models, the Animal2Vec model \cite{schafer2024animal2vec} was developed for meerkat vocalizations. This method uses a self-distillation method to reconstruct masked embeddings, which is similar to the Data2Vec framework. The architecture starts with a feature extractor that is constructed of multiple SincNet-style filterbanks followed by a stack of 1D CNN layers. This is a promising method for UATR, as previous research has shown that decomposition-based features perform effectively in underwater acoustic classification tasks \cite{hummel2024survey}. On top of the feature extractor, the Data2Vec2.0 framework is used, and the complete model is trained in a self-supervised manner. Although the aforementioned models focus on specific animal groups, other bioacoustic models take advantage of more diverse pre-training datasets that encompass a wide range of species. One such model is Perch \cite{ghani2023global}, a lightweight model based on the EfficientNet network \cite{tan2019efficientnet}. Recently, an updated version of this model was released Perch 2.0 \cite{van2025perch}. Both models are designed to be computationally efficient while maintaining strong generalization performance. The authors of Perch 2.0 emphasize that the model can effectively adapt to multiple domains through linear probing, making it a promising candidate for transfer learning in ship type classification. 

Finally, the \textit{Animal Vocalization Encoder based on Self-Supervision} (AVES) \cite{hagiwara2023aves} was included for evaluation. As stated in the name of the model, AVES is optimized using SSL on a broad range of animal vocalizations. The model adopts the HuBERT architecture and training paradigm, extending its application from speech to the bioacoustic domain.



\subsubsection{Marine life sound models}
For the marine life sound models, two models were selected: Google Whale \cite{googlewhale} and Surfperch \cite{williams2025using}. Both models are based on the EfficientNet architecture \cite{tan2019efficientnet} and are trained in a supervised manner.  The Google Whale has been trained with data derived from the NOAA and manually annotated these data to construct a labeled dataset. Their dataset comprises 12 classes, each representing a different type of whale call. The resulting model is designed to detect and classify these calls from PAM recordings. The SurfPerch model \cite{williams2024leveraging} focuses on monitoring the biodiversity within coral reefs. This model is trained on various underwater sound recordings made within coral reefs in a supervised manner. Both models are relatively lightweight and optimized for efficient inference, making them promising candidates for transfer learning in underwater acoustic classification tasks such as UATR.

\subsubsection{Baseline}
\textcolor{black}{The performance of the pretrained audio models were compared to a relatively simple benchmark. This comparison was made to detect whether these complex models were able to outperform a relatively simple and basic method.} This benchmark employs a logistic regression using log Mel-spectrograms as input features for classification. These Mel-spectrograms were constructed using 128 mel bins, a Fast Fourier size of 1,024, and a hop length of 512. Subsequently, mean pooling was applied over the temporal dimension to obtain a fixed-length feature vector for each audio clip. These representations were then used to perform the model performance evaluation.

\subsection{Evaluation set-up}
For the evaluation of the models, the embedding spaces were evaluated using both unsupervised evaluation methods and a supervised evaluation method. During the unsupervised evaluations, the embeddings were clustered and the similarity of the embeddings were measured. Next, during the supervised evaluation the embeddings are utilized to perform ship type classification. 




\subsubsection{Unsupervised: Clustering with Normalized Mutual Information}
First, the embedding space was evaluated using a clustering-based approach. Following the methodology described in \cite{miron2025matters} and \cite{kather2025clustering}, the embeddings are clustered using $K$-Means. The number of training clusters was set equal to the number of classes within the benchmark dataset. In this case, four classes for the Deepship embeddings and five classes for the ShipsEar embeddings. The similarity between the clusters and the ground truth labels was quantified using the Normalized Mutual Information (NMI) score. The score is given by:

\begin{equation}\label{eq:NMI}
    NMI = \frac{2 \times I(Y;C)}{[H(Y) + H(C)]},
\end{equation}
where $Y$ corresponds to the class label, $C$ to the cluster label, function $H(.)$ to the entropy and $I(Y;C)$ to the mutual information between the cluster labels and the class labels. This score ranges between 0, indicating no mutual information, and 1, indicating a perfect correlation.


\subsubsection{Unsupervised: Similarity scores with ranking}
In addition to the clustering evaluation, the similarity between neighboring data samples in the test set was analyzed. Each test sample is treated as a query, and the cosine similarity score is calculated on all other test samples, as described in \cite{schwinger2025foundation} and \cite{van2025perch}. The samples were then ranked from most similar to least similar. From this ranking, the ROC-AUC score is calculated, where samples belonging to the same class as the query were considered positive, and all others were treated as negatives.

\subsubsection{Supervised: Ship type classification}
To finalize the evaluation, automated UATR was introduced as a down-stream task. The performance for ship-type classification was measured by freezing the backbone of each pretrained audio model and extracting the embeddings. Then, a linear probe was trained using these embeddings to predict the type of ships. This approach eliminates the need to fine-tune the entire model, which reduces the number of trainable parameters and minimizes the amount of labeled data required for training \cite{van2025perch}.

\section{Results}
This section presents the results of both the unsupervised and supervised evaluation methods across the various pretrained models. The section is concluded with a comparison of the performance of the HuBERT pretrained in various domains. 


\subsection{Unsupervised evaluation}
First, the embeddings are evaluated in an unsupervised manner. During this evaluation, the geometric separability of the complete embedding space is evaluated by comparing this separability with the label information. This separability is defined by the clusters optimized by a K-Means algorithm or by defining neighboring datasamples by the cosine similarity.

\subsubsection{Clustering with Normalized Mutual Information}
This evaluation assesses whether the clusters defined by a simple $K$-Means algorithm contain informative structure with respect to the ground-truth labels of the benchmark datasets. As shown in Table \ref{tab:NMI_values}, most models achieve low NMI scores that are comparable to the performance of the baseline model. Unlike the classification evaluation, where most of the models outperformed the baseline, the clustering evaluation does not show this. The best performing model in this evaluation for ShipsEar is AvesEcho, achieving an NMI of 0.39. For Deepship, the best performing model is Data2Vec, which reaches an NMI score of 0.18. Interestingly, Data2Vec and Wav2Vec2.0 exhibit an opposite trend. Here, the accuracy scores of Data2Vec were lower compared to Wav2Vec2.0 in the classification evaluation. However, in the clustering evaluation, Data2Vec outperforms Wav2Vec2.0. Overall, these findings indicate that the clusters produced by the $K$-Means algorithm do not appear to align well with the true ship-type classes, suggesting limited semantic separability in the embedding spaces.

\begin{table}[ht]
    \centering
    \begin{tabular}{l|l|c|c}
        Domain & Model & Deepship & ShipsEar \\
        \hline
         \multirow{ 3}{*}{General Audio} & AudioMAE & 0.10 & 0.18 \\
         & BEATS &  \underline{0.13} & 0.22 \\
         & HuBERT AS & 0.10 & \underline{0.23} \\
         \hline
         \multirow{ 4}{*}{Speech} & Wav2Vec2.0 & 0.06 & 0.13 \\
         & Data2Vec & \textbf{0.18} & \underline{0.24} \\ 
         & WavLM & 0.04 & 0.17 \\
         & HuBERT & 0.04 & 0.15 \\
         \hline
         \multirow{ 7}{*}{Bioacoustics} & BirdMAE & \underline{0.14} & 0.26 \\ 
         & BirdNet & 0.05 & 0.30 \\
         & AvesEcho & 0.15 & \textbf{0.39} \\
         & Animal2Vec MK & 0.03 & 0.14 \\
         & Perch & 0.10 & 0.15 \\ 
         & Perch 2.0 & \underline{0.14} & 0.28 \\
         & AVES & 0.09 & 0.19 \\
         \hline
         \multirow{ 2}{*}{Marine life sounds} & Google Whale & 0.06 & \underline{0.16} \\
         & SurfPerch & \underline{0.10} & \underline{0.16 }\\
         \hline
         Baseline & Mel Spectrogram & 0.00 & 0.16 \\
         \hline

    \end{tabular}
    \caption{NMI values of cluster performance on benchmark datasets. The overall best performance in bold and the best performance per domain are underlined.}
    \label{tab:NMI_values}
\end{table}

\subsubsection{Similarity scores with ranking}
The next analysis of the embedding space evaluates whether neighboring datasamples within the embedding space are actually derived from the same ship-type. At first glance, the results in Table \ref{tab:similarity} show that none of the pretrained models substantially outperform the baseline model. Although BEATS is the strongest performer among the pretrained models reaching a ROC-score of 0.61 on Deepship and 0.72 on ShipsEar, it performs similarly to the baseline model reaching 0.62 on Deepship and 0.71 on ShipsEar. These results align with the clustering evaluation, suggesting that the neighboring vectors in the embedding space are often not of the same ship type. In other words, the embeddings do not appear to encode consistent or meaningful characteristics of the ship sound, indicating that the geometric separability of the complete embedding space is not primarily driven by ship-type characteristics.




\begin{table}[ht]
    \centering
    \begin{tabular}{l|l|c|c}
        Domain & Model & Deepship & ShipsEar \\
        \hline
         \multirow{ 3}{*}{General Audio} & AudioMAE & 0.60 & 0.63 \\
         & BEATS & \underline{0.61} & \textbf{0.72} \\
         & HuBERT AS & 0.59 & 0.67 \\
         \hline
         \multirow{ 4}{*}{Speech} & Wav2Vec2.0 & 0.54 & 0.55 \\
         & Data2Vec & \underline{0.60} & \underline{0.64} \\ 
         & WavLM & 0.52 & 0.62 \\
         & HuBERT & 0.53 & 0.61\\
         \hline
         \multirow{ 7}{*}{Bioacoustics} & BirdMAE & \underline{0.61} & 0.70 \\ 
         & BirdNet & 0.59 & 0.67 \\
         & AvesEcho & 0.58 & \underline{0.71} \\
         & Animal2Vec MK & 0.57 & 0.64 \\
         & Perch & 0.59 & 0.66 \\ 
         & Perch 2.0 & 0.60 & 0.69 \\
         & AVES & 0.60 & 0.67 \\
         \hline
         \multirow{ 2}{*}{Marine life sounds} & Google Whale & 0.55 & 0.62 \\
         & SurfPerch & \underline{0.59} & \underline{0.66} \\
         \hline
         Baseline & Mel Spectrogram & \textbf{0.62} & 0.71 \\
         \hline

    \end{tabular}
    \caption{Mean ROC-AUC values from ranking based on the cosine similarity. The overall best performance is in bold, and the best performance per domain are underlined.}
    \label{tab:similarity}
\end{table}

\subsubsection{Are the embeddings defined by record ID information?}

\begin{figure}
    \centering
    \includegraphics[width=0.9\linewidth]{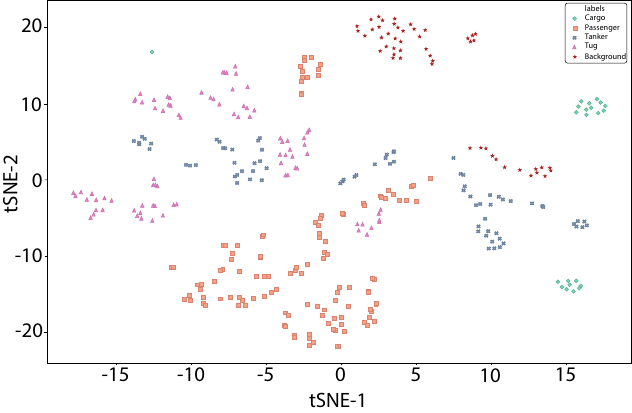}
    \caption{t-SNE plot of the embeddings generated by BEATS using the ShipsEar test set.}
    \label{fig:tSNE_BEATS}
\end{figure}

As can be seen from the results, the BEATS model achieves the best performance or a competitive performance in both evaluation metrics.
To further investigate this, the embeddings of the ShipsEar test set generated by BEATS is visualized in Figure \ref{fig:tSNE_BEATS}. The plots show a clear separation between the classes. However, it can also be seen that the classes form small subclusters within the embedding space. This is especially visible for the background class. In particular, the background class splits into three distinct subgroups, while the test set also contains three individual recordings. This suggests that the geometric separability of the embedding space generated by the pretrained model is mainly driven by individual recording information. 

To determine whether the pretrained BEATS embedding space learned to differentiate recordings, the evaluation was repeated using record IDs instead of ship-type labels. Since individual recordings do not appear in both the training set and the test set, the classification of the recording ID is excluded from this evaluation. For cluster evaluation, the number of clusters for the $K$-Means algorithm was set to the number of individual recordings in the test set. For ShipsEar, this was set to 21, and for Deepship to 237. Table \ref{tab:recordingID} shows that both the resulting NMI and ROC-AUC scores are substantially higher compared to those obtained with label-based evaluation. In particular, the ShipsEar NMI increased from 0.22 to 0.66, while the ROC-AUC increased from 0.72 to 0.93. This shows that the placement of datasamples within the complete embedding space is mainly driven by record ID information. This analysis has been repeated for the other pretrained models, resulting in the same phenomenon. 

\begin{table}[]
    \centering
    \begin{tabular}{l|c|c}
        Metric & Deepship & ShipsEar \\
        \hline
        NMI & 0.62 & 0.66 \\
        ROC-AUC & 0.90 & 0.93\\
        \hline
    \end{tabular}
    \caption{Evaluation of the recording ID information in the BEATS embeddings.}
    \label{tab:recordingID}
\end{table}

\subsection{Supervised evaluation}
To finalize the evaluation, the embedding spaces were evaluated by introducing a downstream task. In this study, this task was defined as classifying various ship-types from two distinct benchmark datasets. This analysis checks whether the pretrained model is able to translate to a newly seen task by training a classifier in a supervised manner for a task that the model was initially not pretrained on.

\subsubsection{Ship type classification: linear probing}
By applying a linear probe, the majority of the models substantially outperform the baseline model (see Table \ref{tab:acc_values}). This indicates that, even though the geometric separability of the embedding space did not indicate containing any ship-type characteristics, a low-cost linear probe was still able to effectively differentiate between various ship-types. 

In general, the results in Table \ref{tab:acc_values} show that the accuracy scores in the ShipsEar dataset are higher compared to those of the Deepship dataset in most cases. This difference was expected because of the way the datasets were split into a training set and a test set. For Deepship, a time-wise split was defined to differentiate between the training set and the test set. This ensured no data leakage, but introduced previously unseen ships in the test set. In addition, seasonality differences can also induce variability between the recordings in the training set and the test set. Together, this explains the overall lower performance on Deepship compared to ShipsEar. 

When examining the performance of individual models, Wav2Vec2.0 has the best performance in the ShipsEar dataset reaching 78\% accuracy. Additionally, BEATS performed the best in the Deepship dataset reaching 65.4\% accuracy, and is also the second best performer in ShipsEar, reaching 74\% accuracy. In contrast, WavLM underperformed relative to the baseline model in both ShipsEar and Deepship. This could be explained by the denoising objective that is used for pretraining, which may suppress relevant noise characteristics of the ship, and thus remove critical information required for accurate classification. Another notable result is the high performance of Animal2Vec, despite its use of the lowest sampling rate (8 kHz) among all models. This suggests that this lower temporal resolution did not affect the model's ability to distinguish between ship types. In contrast, marine life sound models perform poorly. Although both models were optimized for various underwater sounds, they failed to make the translation to ship noises. Their supervised training in marine life sounds likely limited their ability to generalize to other acoustic domains such as ships. Similarly, BirdNET, Perch, and Perch2.0 were also pretrained in a supervised manner. However, these models benefited from multiple labeled datasets, introducing more varied data diversity, and consequently improved their generalization capability. In addition to linear probing, a more advanced probing method, attentive probing, has also been evaluated. Overall, these results did not compete with the linear probe, making the linear probe the best performer while also being computationally more efficient.

\begin{table}[ht]
    \centering
    \begin{tabular}{l|l|c|c}
        Domain & Model & Deepship & ShipsEar \\
        \hline
         \multirow{ 3}{*}{General Audio} & AudioMAE & 61.9\% & 67.2\% \\
         & BEATS & \textbf{65.4\%} & \underline{74.0\%}  \\
         & HuBERT AS & 54.2\% & 42.7\% \\
         \hline
         \multirow{ 4}{*}{Speech} & Wav2Vec2.0 & \underline{56.8\%} & \textbf{78.0\%} \\
         & Data2Vec & \underline{56.8\%} & 53.1\% \\ 
         & WavLM & 49.7\% & 45.0\%\\
         & HuBERT & 50.1\% & 32.8\% \\
         \hline
         \multirow{ 7}{*}{Bioacoustics} & BirdMAE & 63.4\% & 67.2\%  \\ 
         & BirdNet & 58.6\% & 57.5\%\\
         & AvesEcho & 57.22\% & 64.26\% \\
         & Animal2Vec MK & \underline{64.0\%} & 61.0\%\\
         & Perch & 55.0\% & 64.3\% \\
         & Perch 2.0 & 61.4\% & \underline{73.2\%} \\
         & AVES & 56.4\% & 59.1\% \\
         \hline
         \multirow{ 2}{*}{Marine life sounds} & Google Whale & 45.9\% & 50.6\%\\
         & SurfPerch & \underline{57.8\% }& \underline{58.9\%} \\
         \hline
         Baseline & Logistic Regression & 56.4\% & 48.6\% \\
         \hline

    \end{tabular}
    \caption{Accuracy values of performance on benchmark datasets using linear probing. The overall best performance is in bold, and the best performance per domain are underlined.}
    \label{tab:acc_values}
\end{table}

\subsubsection{Is the classification performance driven by record ID information?}

To assess whether the observed classification performance is primarily driven by recording-specific artifacts rather than ship-type information, a label-shuffling control experiment was conducted on the BEATS embeddings. Specifically, the ship-type labels were randomly permuted  across recordings and retrained the linear probe on these shuffled labels. If the embeddings mainly encoded recording identity or other spurious cues that correlate with ship type, the classifier would still achieve relatively high accuracy even after shuffling. Instead, performance dropped sharply to 44.6\% accuracy on ShipsEar and 22.9\% on Deepship. This decrease in performance indicates that once the semantic link between embeddings and ship types is broken, the model cannot generalize meaningfully. In other words, the original classification performance relies on genuine ship-type structure present in the embedding space, not solely on recording-specific information.

To further probe this observation, the clustering analysis was repeated using the logits of the trained linear probe rather than the raw embeddings generated by BEATS. Clustering in logit space yields substantially higher alignment with ship types, with NMI scores of 0.42 for ShipsEar and 0.36 for DeepShip, approximately twice the NMI obtained from the full embedding space. This suggests that while the embedding space contains mixed factors of variation, the linear probe successfully isolates ship-type–relevant dimensions, suppressing recording-specific effects.

To verify that this result is not an artifact of dimensionality reduction, the analysis is repeated on the reduced embedding space of BEATS by performing a \textit{Principal Component Analysis} (PCA). The reduced embedding space resulted in a slightly higher NMI score from 0.13 in Deepship and 0.22 on ShipsEar originally, to 0.14 on Deepship and 0.27 on ShipsEar. This shows that indeed the increase in the NMI score on the logits is not driven by the dimensionality reduction, but rather by the ship characteristic information. 

\begin{figure}
    \centering
    \includegraphics[width=0.9\linewidth]{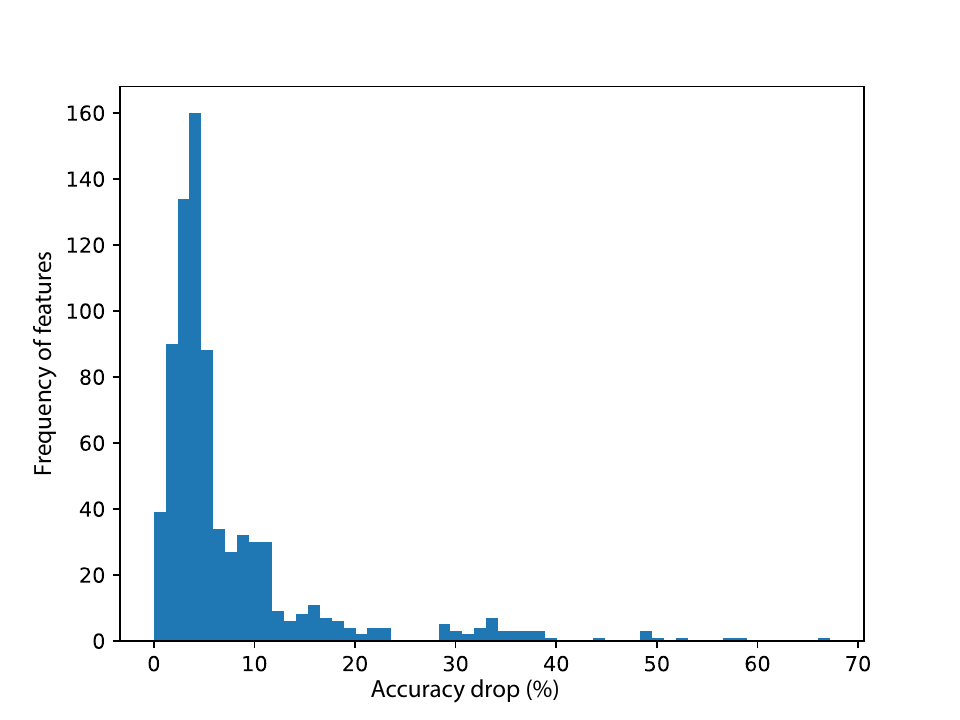}
    \caption{Histogram of the accuracy drop of removing each feature from the BEATS embeddings individually on the classification of ShipsEar.}
    \label{fig:FeatureImportance}
\end{figure}

The same experiments have been conducted on the other pretrained audio embeddings, showing similar results. All together, these results suggest that the linear probe can select a small subset of the complete embedding space to extract ship type characteristics even though the complete embedding space is overruled by record ID information. To support this claim, a feature importance analysis was performed on the original BEATS classifier. During this analysis, each feature is set to 0 at the time while the original classifier tries to classify the various ship types. The results in Figure \ref{fig:FeatureImportance} show a histogram of the drop in accuracy per feature. Here, the figure shows that the majority of the elimination of the features did not influence the performance of the model. Only a small portion of the features had a severe impact on the performance of the model, underscoring the hypothesis that the linear layer selects a small subset of the embedding space to perform satisfactory classification.

\subsection{Influence on data domain: Comparison of HuBERT}
\begin{figure*}[t!]
    \centering
    \begin{subfigure}[t]{0.5\textwidth}
        \centering
        \includegraphics[width=0.9\linewidth]{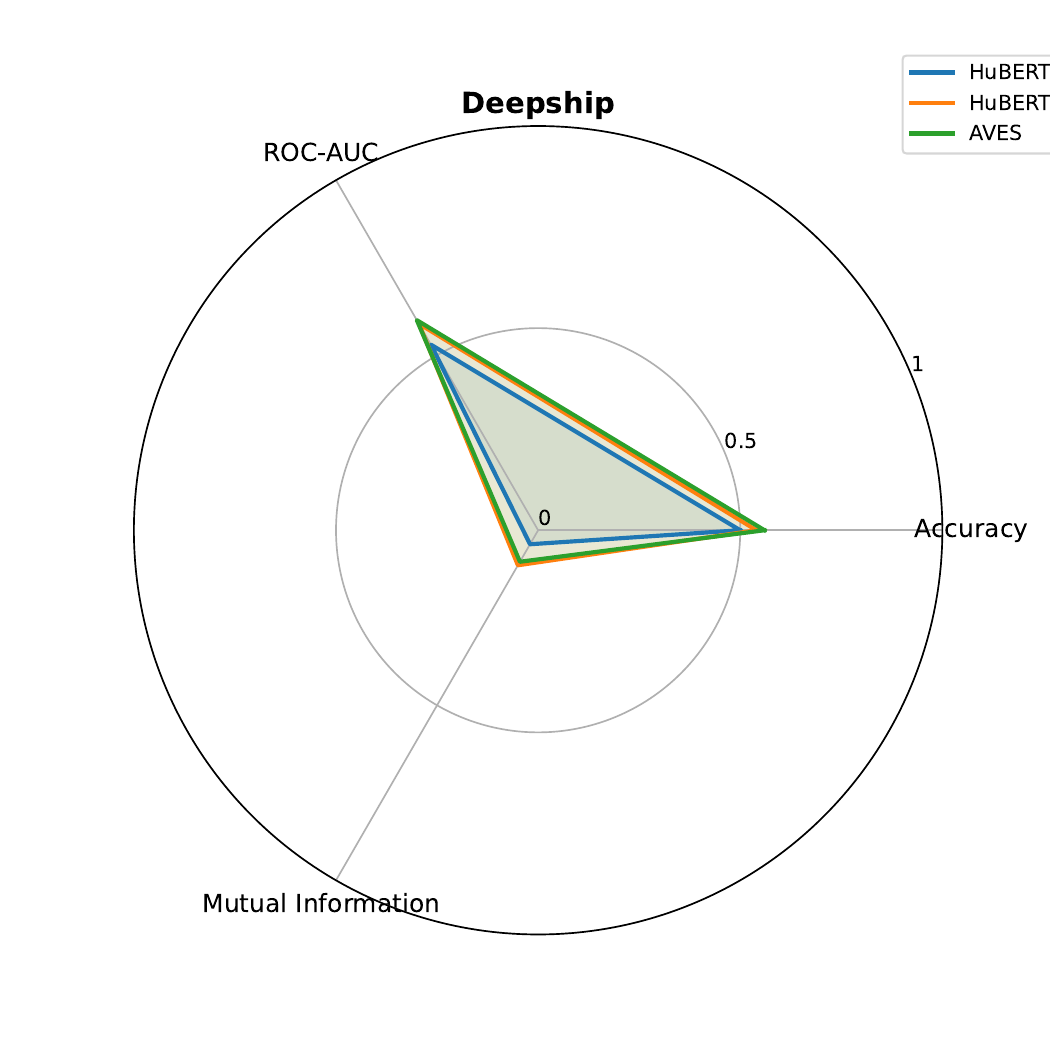}
        \caption{Performance on Deepship}
    \end{subfigure}%
    ~ 
    \begin{subfigure}[t]{0.5\textwidth}
        \centering
        \includegraphics[width=0.9\linewidth]{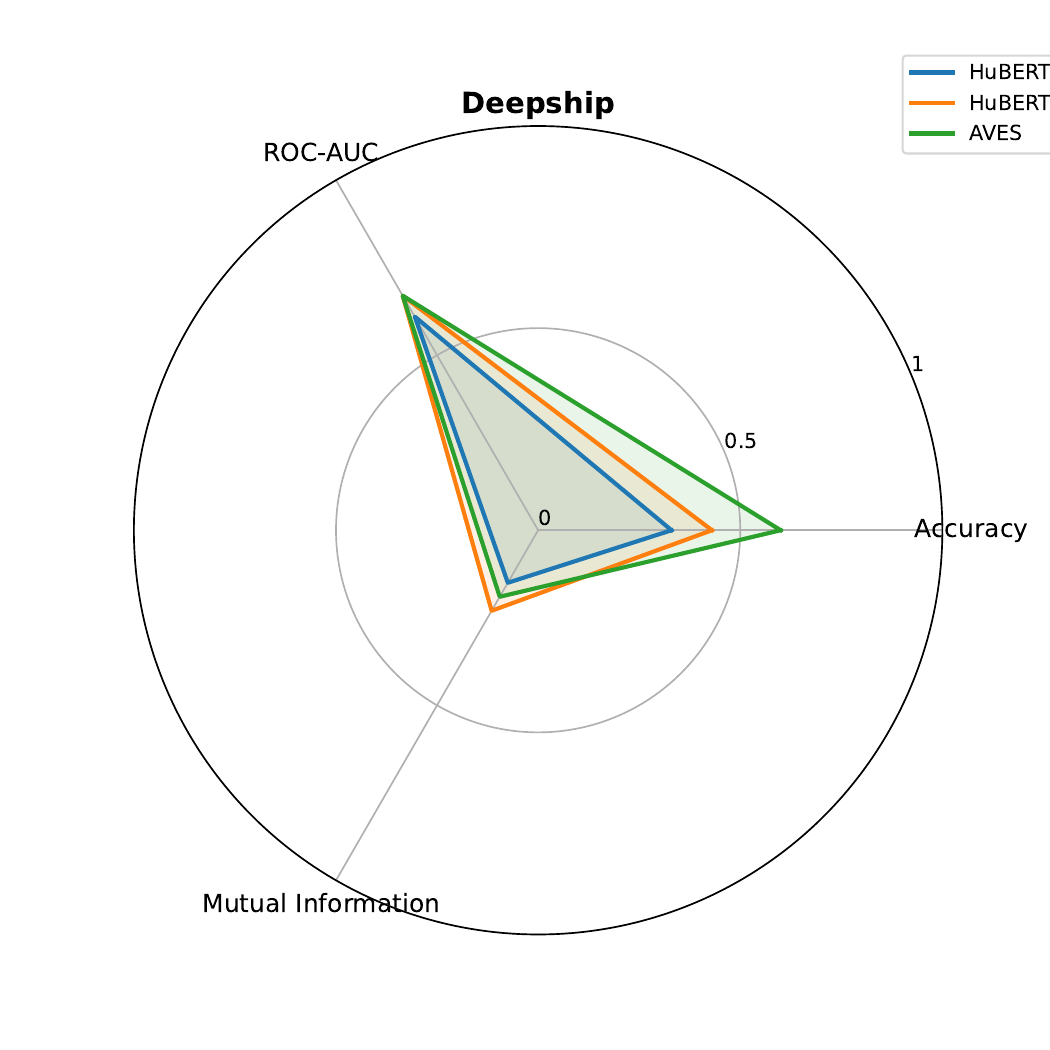}
        \caption{Performance on ShipsEar}
    \end{subfigure}
    \caption{Spider plot of the performances of HuBERT-based models trained on datasets from different domains.}
    \label{fig:SpiderPlot_HuBERT}
\end{figure*}

Among the evaluated models, three models can be directly compared: HuBERT, HuBERT (AS), and AVES. All of these models share the same architecture and training paradigm, but differ in their pretraining dataset. By comparing the performance of these three models, the influence of the pretraining domain can be visualized. The results summarized in Figure \ref{fig:SpiderPlot_HuBERT} show that all three models have limited performance on evaluation tasks, suggesting that the HuBERT pseudo-label-based learning objective may not effectively capture features relevant to the acoustics of the ship. However, some domain-related differences can be seen. The model pretrained in speech is underperforming the other models for all three evaluation methods, indicating a poor transfer from speech to UATR. Interestingly, the bioacoustic model (AVES) even slightly outperforms the general audio model. This phenomenon can also be seen in other models, where the majority of the bioacoustic models slightly outperform the general audio models. Especially the models that have been trained on bird data show good performance, which could be explained by the diversity of bird data. Bird-based datasets, in particular, contain recordings from varied environments and devices worldwide, exposing models to a wide range of natural and anthropogenic sounds. The model may benefit from this diversity and, therefore, be able to generalize slightly more to other domains. 


\section{Discussion}
This work presents an extensive empirical study on the transfer learning of pretrained audio models from various domains to UATR. When comparing performance between different pretrained audio models, it is important to consider variations in preprocessing techniques specific to each model. For example, differences in sample rates may influence the results, where higher sample rates retain more information than lower sample rates, potentially benefiting downstream performance \cite{schwinger2025foundation}. The same principle applies to the embedding size; larger embedding sizes may retain more detailed information, but can also increase the risk of overfitting to the pretraining data \cite{hummel2025computation}. Nevertheless, this variation enables the examination of model diversity, providing deeper insight into how design choices impact downstream performance.

Initially, the geometric separability of the embedding space was evaluated by unsupervised metrics. Both metrics showed a lower or comparable performance compared to the baseline model. This suggests that the geometric separability of the pretrained embeddings was similar to the baseline, despite outperforming the baseline in the classification evaluation. In other words, embeddings that were close in the embedding space were not necessarily derived from the same ship class. To explore this, the embedding spaces were analyzed in more detail. Visualization of BEATS embeddings indicated that embeddings from the same recording clustered closely together, and similarity scores were largely driven by recording-specific properties.  
Environmental factors, hydrophone types, and recording configurations likely introduce greater variance between the recordings than the various ship-types.

Despite this, the performance of the pretrained models was evaluated using linear probing to classify ship types from the labeled benchmark datasets. At first glance, most of these models achieved satisfactory accuracy scores, given that the ship-type classification is a multi-class problem. Surprisingly, the models optimized for marine life showed poor performance. This was unexpected, as these models were presumed to have learned some general underwater acoustic properties. Other research has similarly reported that the generalizability of GoogleWhale to other related tasks is limited \cite{kather2025clustering}. In contrast, models pretrained on bioacoustic or general audio data demonstrated stronger transfer performance to UATR. Among them, BEATS achieved particularly strong results in both ShipsEar and DeepShip datasets. Its robustness in DeepShip is especially notable, given that the test set was constructed using a time-based split, which means that the majority of individual ships in the test data were absent from the training set \cite{hummel2025computation}. This finding indicates that BEATS was able to extract meaningful ship-type characteristics directly from raw audio that generalize to previously unseen vessels. 



Overall, these results indicate that, while ship-type information is linearly decodable from the embeddings, the global structure of the embedding space is dominated by recording-specific characteristics. This suggests that recording artifacts and session-dependent properties exert a stronger influence on the learned representations than ship-type semantics. The linear probe effectively learned to downweight embedding dimensions dominated by recording-specific information, focusing instead on the small subset that carried ship-identifying features. This acts much like a feature selection mechanism, highlighting that even when the embedding space is noisy and entangled, simple models can extract meaningful signals. However, supervision is needed to do a meaningful selection for the downstream task. These insights underline both the promise and limitations of pretrained audio models in transferring to complex underwater acoustic tasks like UATR.



\section{Conclusion}
\textcolor{black}{In conclusion, pretrained audio models effectively transfer to UATR tasks. Especially audio models pretrained on general audio tasks were able to transfer successfully to ship sounds, where the BEATS model outperformed the other models. An in depth analysis of the BEATS embedding space showed that the embeddings were mainly driven by recording specific characteristics. However, a low cost linear probe was able to extract a small subset of the embedding space to accurately classify ship types in two different benchmark datasets. This means that pretrained audio models can effectively be reused for UATR tasks by only optimizing a low cost linear probe. This greatly reduces the need for a large scale labeled dataset.} This may enable pretrained audio models to transfer to other related underwater acoustic applications, such as climate change measurements and marine life tracking. Together, this study gives the potential for automated exploration of the ocean ecosystem without the need for large-scale annotations and heavy computations.

\hfill

{\bf Declaration of Competing Interest:} None of the authors has any relationship with people or organizations that can influence their work.

\bibliographystyle{elsarticle-num} 
\bibliography{references}

\end{document}